\documentclass[10pt,twocolumn,letterpaper]{article}

\usepackage{wacv}              

\usepackage{multirow}
\usepackage{graphicx}
\usepackage{amsmath}
\usepackage{amssymb}
\usepackage{booktabs}
\usepackage{subcaption}
\usepackage{colortbl}
\usepackage[table]{xcolor}
\usepackage{algorithm}
\usepackage[noend]{algpseudocode}

\usepackage[pagebackref,breaklinks,colorlinks]{hyperref}

\usepackage[capitalize]{cleveref}
\crefname{section}{Sec.}{Secs.}
\Crefname{section}{Section}{Sections}
\Crefname{table}{Table}{Tables}
\crefname{table}{Tab.}{Tabs.}

\def\wacvPaperID{07} 
\def\confName{WACV}
\def\confYear{2024}

\begin{document}

\title{Towards On-device Learning on the Edge: \\ Ways to Select Neurons to Update under a Budget Constraint}

\author{Aël Quélennec
\qquad
Enzo Tartaglione
\qquad
Pavlo Mozharovskyi
\qquad
Van-Tam Nguyen\\
LTCI, Télécom Paris, Institut Polytechnique de Paris\\
19 Place Marguerite Perey, 91120 Palaiseau, France\\
{\tt\small \{name.surname\}@telecom-paris.fr}}

\maketitle

\begin{abstract}
In the realm of efficient on-device learning under extreme memory and computation constraints, a significant gap in successful approaches persists. 
Although considerable effort has been devoted to efficient inference, the main obstacle to efficient learning is the prohibitive cost of backpropagation. 
The resources required to compute gradients and update network parameters often exceed the limits of tightly constrained memory budgets. This paper challenges conventional wisdom and proposes a series of experiments that reveal the existence of superior sub-networks. Furthermore, we hint at the potential for substantial gains through a dynamic neuron selection strategy when fine-tuning a target task. Our efforts extend to the adaptation of a recent dynamic neuron selection strategy pioneered by Bragagnolo et al. (NEq), revealing its effectiveness in the most stringent scenarios. Our experiments demonstrate, in the average case, the superiority of a NEq-inspired approach over a random selection. This observation prompts a compelling avenue for further exploration in the area, highlighting the opportunity to design a new class of algorithms designed to facilitate parameter update selection. Our findings usher in a new era of possibilities in the field of on-device learning under extreme constraints and encourage the pursuit of innovative strategies for efficient, resource-friendly model fine-tuning.
\end{abstract}


\section{Introduction}

\begin{figure}[t]
    \centering
     \includegraphics[width=\columnwidth]{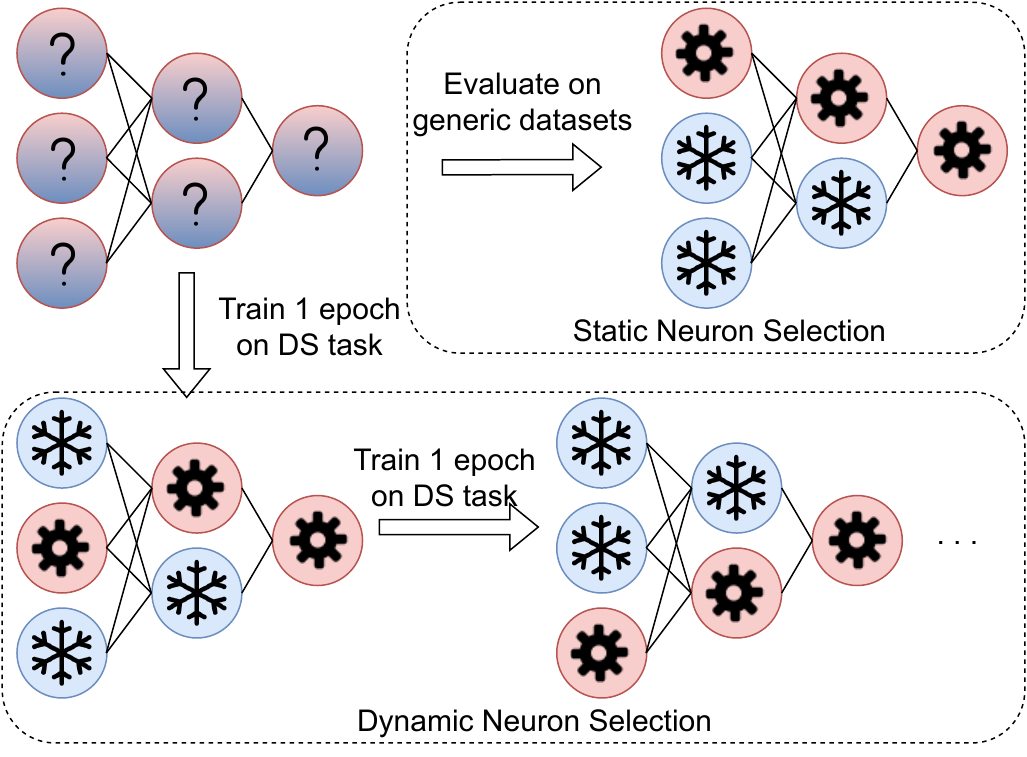}
    \caption{Comparison between static neuron update selection (up-right corner) and dynamic selection (down). While the static strategy evaluates the neurons to update on generic datasets and the sub-network will remain static for the whole training on the down-stream (DS) task, with the dynamic selection it can change after every epoch.}\vspace{-2ex}
    \label{fig:teaser}
\end{figure}

In recent years, the dynamic landscape of deep learning has witnessed remarkable progress across a multitude of domains, spanning from computer vision~\cite{Mo_2023_WACV,app12105178} and speech recognition~\cite{Burchi_2023_WACV} to natural language processing~\cite{wu2023graph}. This evolution, coupled with the escalating prowess of novel model architectures, has firmly entrenched deep learning as a pivotal technological force. The accelerated growth in hardware capabilities, alongside the advent of AI-dedicated processing units, has enabled the training of increasingly expansive models, as indicated by prior work~\cite{8421507}. Yet, the intrinsic demand for substantial computational resources during training, and the consequential surge in energy consumption~\cite{strubell2019energy}, underscore a pressing challenge. Whether it is at inference or training time, the inextricable interplay between three vital factors—computational resource utilization, energy efficiency, and inference time—requires meticulous attention. The mission at hand is twofold: to empower smaller institutions to partake in the development and training of state-of-the-art models and to conscientiously curtail the carbon footprint of the deep learning community.\\

On-device training stands as a pivotal advancement in the realm of artificial intelligence, carrying profound implications across various applications~\cite{incel2023device}. One of its primary merits is its ability to facilitate continuous model improvement even after deployment (when combined with online learning strategies)~\cite{hayes2022online}. This becomes especially beneficial when the nature of the data evolves or when user interactions are integral to the application. Language models, for instance, can adapt to ever-changing linguistic trends, colloquialisms, and user preferences, resulting in more accurate and relevant predictions. Moreover, on-device training empowers models with lifelong learning capabilities: this means they can accumulate knowledge and adapt as they encounter new data, making them invaluable in fields such as healthcare where medical knowledge is continually evolving~\cite{poquet2021developing}. On-device training also ushers in the era of user customization, allowing models to adapt to individual preferences and writing styles, thus enhancing the user experience.\\

Although making models efficient at inference is a well-known and even currently explored challenge, efficient on-device learning under extreme memory and computation budget is a relatively new one. More specifically, compared to inference, the biggest obstacle resides in the cost of performing backpropagation (BP) (computing the gradient and updating all the parameters in the network quickly exceeds tight memory budgets). Although some approaches to perform learning on-device are alternatives and modifications to BP such as unsupervised learning for image segmentation~\cite{yang2023device} or the recently proposed Forward-Forward algorithm~\cite{hinton2022forward} and PEPITA~\cite{pau2023suitability}, they are currently below the performance achieved by BP. Building on this, Lin~\emph{et~al.} was finally able to successfully fine-tune a deep model with an extreme memory budget (under 256kB)~\cite{lin2022device} by statically determining which parts of a pre-trained model should be updated on any downstream task. The authors here implicitly whisper that there is essentially a sub-network providing sufficiently general features to be adapted to most of the downstream tasks, and can be kept frozen while updating a minor part of it. This hypothesis can be also seen, in a certain sense, as an extension of the lottery ticket hypothesis~\cite{frankle2018the} applied on the BP graph in isolation~\cite{tartaglione2022rise}, where just a sub-network needs to be updated to find a target performance.  

In this paper, we challenge the vision by Lin~\emph{et~al.} under extreme memory budget constraints, proposing some experiments where we observe that better sub-networks exist. We summarize our contributions as follows.
\begin{itemize}
    \item We modify a dynamic neuron update selection strategy, NEq~\cite{bragagnolo2022update}, to work under extreme memory budgets (Sec.~\ref{sec:NEQ++}).
    \item We compare the static selection strategy~\cite{lin2022device} to our dynamic approach (Fig.~\ref{fig:teaser}), reporting in the average case the superiority of the dynamic one (Sec.~\ref{sec:dynvsstatic}).
    \item We introduce a random dynamic neuron selection baseline, where neurons to update are chosen randomly. Our proposed strategy, in the majority of the tested setups, shows its superiority (Sec.~\ref{sec:dynamicvsrandom}). Such observation opens the road to more study in the field, where a new class of algorithms for parameter update selection should be designed.
\end{itemize}

\section{Related work}
\label{sec:sota}

\noindent\textbf{On-device learning} On-device learning is a growing field of research due to the increasing number of embedded devices for IoT applications. To this day, the principal modus operandi is to train a model offline and then compress and deploy it on-device for inference only. However, such methodology often yields poor performance due to real data distribution shifts from training data~\cite{spadaro2023shannon}. Research in continual learning shows that it is an efficient solution to adapt models to distribution shifts in post-deployment scenarios. Continual learning mimics human behavior in sequentially acquiring and retaining knowledge across various tasks~\cite{ContinualLifeLong, zenke2017continual}. Evidently, in such a scenery catastrophic forgetting, which is the undesired loss of information acquired from previous tasks, is massively addressed by the research community~\cite{CFinterference, ContinualLifeLong, Kirkpatrick_2017} defined as the significant loss of earlier acquired knowledge during the learning of new tasks. However, embedded devices are highly constrained in computational and memory resources while training: especially backpropagation is very costly, making it a true challenge to achieve without significantly affecting model performance. For on-device learning based on backpropagation, we can distinguish two types of approaches, often combined to attain the best training accuracies under resource constraints: improving the architecture's efficiency and performing sparse updates. 

\noindent\textbf{Efficient architectures.} The first is to design resource-efficient neural networks such as MobileNets~\cite{sandler2018mobilenetv2}, EfficientNets~\cite{tan2019efficientnet} or MCUNet~\cite{lin2020mcunet}, or even some efficient versions of transformers~\cite{elliott2021tiny}. From a certain perspective, this method consists of the reduction of trainable parameters, impacting directly on memory and computation reduction. This method is also often paired to quantization to reduce the memory footprint on-device. More general approaches in such perspective involve automatic Neural Architecture Search (NAS)~\cite{li2020random, hardt2016equality, liu2018darts}. NAS automates the exploration of neural network architectures, optimizing factors like model size, performance in terms of accuracy, and required FLOPs, often employing multi-objective optimization. Evidently, despite the big effort in making these strategies as efficient as possible~\cite{jin2019auto,yang2020cars,pham2018efficient}, fine-tuning a pre-trained, off-the-shelf architecture remains the least energy-consuming approach.

\noindent\textbf{Sparsely update the model.} The second method is to sparsely update the network. As shown in~\cite{cai2020tinytl}, the memory footprint of backpropagation is greatly due to the loading of each layer's input tensor from which the gradient is computed. However, in the context of fine-tuning a pre-trained model given a shifted distribution, it is better to surgically select a subset of layers to train while freezing the rest of the network~\cite{lee2022surgical}, leading to drastic savings regarding activation costs while achieving good performances. Such a concept is what led Lin~\emph{et~al.} to design the Sparse Update (SU), an optimized static selection of a subset of layers to train while on-device~\cite{lin2022device}. However, the SU configurations are very costly to find as they require an evolutionary algorithm to iterate over many trainings. In the next section, we will present an approach embodying a different philosophy, challenging a static update graph allocation but rather identifying dynamically the partition of the model better to be updated. 
\section{Sub-network selection under constraint: from static to dynamic}

In this section, we present our novel contributions, which extend upon SU and NEq selection techniques. We discuss the inherent limitations of these methods within the on-device learning domain, starting from an overview of SU schemes, and finally introducing two innovative approaches, inspired by the NEq approach. Our primary objective is to address these limitations by proposing dynamic online neuron selection mechanisms, tailored to accommodate stringent memory budgets.

\subsection{Unboxing Sparse Update}

In \cite{lin2022device}, Lin~\emph{et~al.} present a pioneering algorithm-system co-design framework, enabling on-device training to adapt models to sensor data without compromising privacy. Overcoming the memory limitations of IoT devices, they were indeed able to achieve on-device training with just 256kB of memory. The authors were able to accomplish this thanks to a combination of four elements: 
\begin{enumerate}
    \item SU, corresponding to a selection of a subset of layers and weights to optimize during training, in order to reduce the memory footprint of gradient computation during backpropagation;
    \item the usage of networks specifically conceived for tight resource environments, such as MobileNetV2 or MCUNet;
    \item the introduction of a quantization-aware scaling in order to stabilize quantized gradient update;
    \item the design of ``Tiny Training Engine'', an efficient training system transforming the actual training into slim binary codes.
\end{enumerate}

The SU approach builds on top of the assumption that just a sub-network can be fine-tuned to achieve good performance. To validate this, Lee~\emph{et~al.} demonstrate that when fine-tuning a pre-trained model on a target dataset, specifically selecting a subset of layers to train depending on the type of distribution shift does not affect performance negatively and instead can even improve it~\cite{lee2022surgical}. The reduction of training memory usage through selective layer training during fine-tuning is especially interesting when considering on-device learning where models are typically pre-trained offline on a large dataset. Lin~\emph{et~al.} took this further by analyzing the impact of training the bias at different depths as well as different layer update rates, finding a positive correlation between combined training results and individual accuracy gains. Their methodology is the following: for a classification model pre-trained on ImageNet-1k, the objective is to determine the influence on accuracy gain of training different individual bias and layer configurations on the downstream task of Visual Wake Word classification~\cite{chowdhery2019visual}, in comparison with only training the classifier layer.

\noindent\textbf{Bias update utility.} Lin~\emph{et~al.} conducted a comprehensive quantitative evaluation to assess the impact of bias updates within the network. Specifically, they conducted a series of individual training sessions, where every bias from the $l$-th layer onward (from the $l$-th layer to the output) was updated. Each training iteration resulted in a relative accuracy improvement $\Delta \mathcal{A}_l^{\text{bias}}$ when compared to training only the classifier. Through their empirical analysis, the authors noted a steady increase in accuracy enhancement, eventually reaching a point of saturation at a specific depth within the network.

\noindent\textbf{Layer update utility.} At each network layer, Lin~\emph{et~al.} performed a series of experiments to assess the impact of training with varying channel ratios $\zeta \in \{0, \frac{1}{8}, \frac{1}{4}, \frac{1}{2}, 1\}$ on the relative accuracy gains $\Delta \mathcal{A}_{l, \zeta}^{\text{w}}$. This process entailed performing a set of training sessions equal to the number of network layers, each multiplied by the number of ratios tested.

\noindent\textbf{Cost of individual configurations.} The analysis of memory usage for each layer reveals that early layers of the network have a high activation cost $\mathcal{C}_l^{\text{act}}$ (cost of loading the layer's input tensor to compute the gradient) and a low-weight memory cost $\mathcal{C}_l^{\text{w}}$ (cost of loading the layer weights in memory). Late layers in the network have the opposite behavior and middle layers have both low activation and memory cost.

\noindent\textbf{SU scheme selection.} Given each bias depth and layer ratio relative accuracy gains, Lin~\emph{et~al.} observe that the aggregation of different configurations leads to a final accuracy that is positively correlated with the sum of relative accuracy gains from each individual configuration. They use this empirical observation in an evolutionary algorithm to optimize the final sum of $\{\Delta \mathcal{A}_{l, \zeta}^{\text{w}}, \Delta \mathcal{A}_l^{\text{bias}}\}\forall l, \zeta$, given a memory budget constraint. This evolutionary search results in a specific subset of layers ratios and bias depth to update which is referred to as the SU scheme.

\noindent\textbf{Beyond SU.} The SU approach is labor-intensive and costly, requiring multiple trainings for various layers and update combinations followed by an evolutionary search to find the optimal configuration. Additionally, Lin~\emph{et~al.} compute the relative accuracy gains on one downstream task and then apply the SU scheme found to many different target datasets. As demonstrated theoretically by Lee~\emph{et~al.}~\cite{lee2022surgical} and then validated empirically by Kwon~\emph{et~al.}~\cite{kwon2023tinytrain}, the optimal subset of layers to fine-tune depends on the target dataset, meaning that the contribution analysis should be performed for each downstream task. Additionally, the SU configuration remains unchanged throughout network fine-tuning, leading to the same neurons being updated over and over until potential over-fitting while other neurons remain frozen even though they could require some training to improve performance. This motivated us to move to dynamic update schemes - here follow the preliminaries to set up bases for a dynamic update scheme under a memory (and computation) budget.

\subsection{A dynamic approach: Neurons at Equilibrium}
\begin{figure*}[t]
    \centering
     \includegraphics[width=0.75\textwidth]{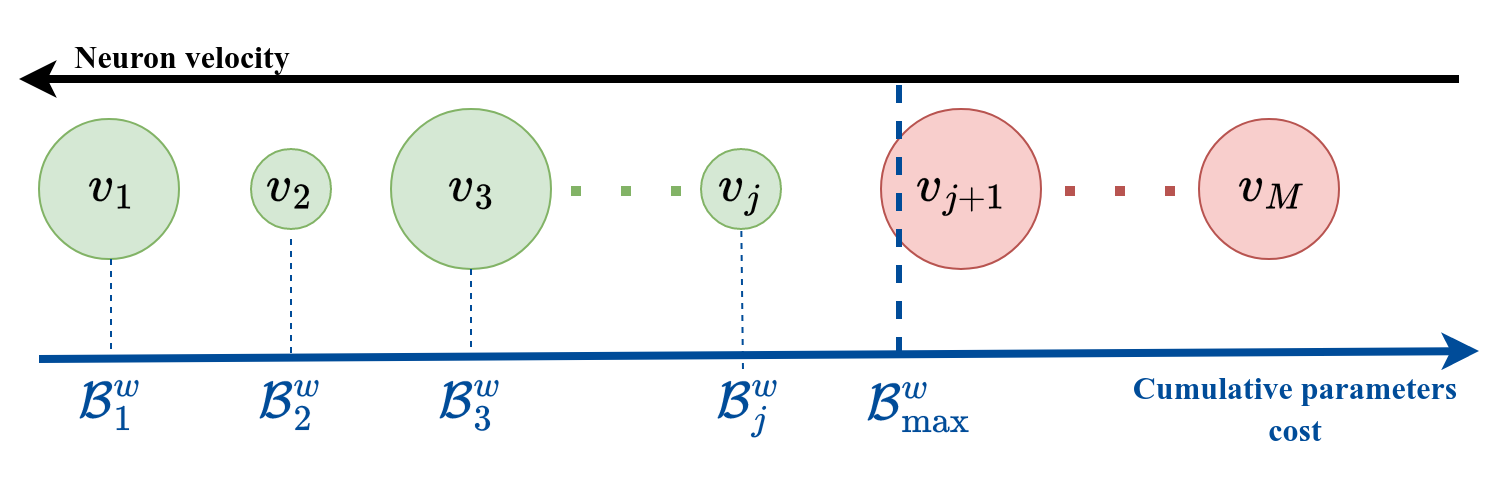}
    \caption{Selection of neurons to update given a network of $M$ neurons and a budget $\mathcal{B}_{\text{max}}^{w}$. The cost $\mathcal{C}_{i}^{w}$ is proportional to the size of the circle which represents the $i$-th neuron. The neurons selected for the update are in green, while those frozen in red.
    } 
    \label{fig:Neuron_selection}
\end{figure*}
In this section, we will present NEq, an algorithm proposed by Bragagnolo~\emph{et~al.} which targets energy consumption reduction at training time with no performance loss~\cite{bragagnolo2022update}. NEq aims at reducing training time and cost without affecting performance by progressively selecting neurons to freeze throughout training.\\
Let us define the output of the $i$-th neuron when the input $\boldsymbol{x}$ is fed to the whole model trained after $t$ epochs as $\boldsymbol{y}_{i,\boldsymbol{x}}^t$. For a given set of inputs $\boldsymbol{x}\in \mathcal{D}_{\text{val}}$ (where $\mathcal{D}_{\text{val}}$ is the validation set), we can compare each $n$-th output $y_{i,\boldsymbol{x},n}^t$ with $y_{i,\boldsymbol{x},n}^{t-1}$ by computing the cosine similarity $\phi_{i}^t$ between all the outputs of the $i$-th neuron at time $t$ and at time $t\!-\!1$ for the whole validation set $\mathcal{D}_{\text{val}}$:
\begin{equation}
    \phi_{i}^t = \sum_{\boldsymbol{x}\in \mathcal{D}_{\text{val}}} \sum_{n=1}^{N_{i, \boldsymbol{x}}} \hat{y}_{i,\boldsymbol{x},n}^{t} \cdot \hat{y}_{i,\boldsymbol{x},n}^{t-1},
\end{equation}
where $N_{i, \boldsymbol{x}}$ is the cardinality of the output for the $i$-th neuron when $\boldsymbol{x}$ is fed as input to the neural network. The variations of $\phi_{i}$ over different epochs inform us about the stability of the relationship between the input and the output of the $i$-th neuron. To quantify the amount of variation between epochs, we compute the relative variation
\begin{equation}
    \Delta \phi_i^t = \phi_{i}^{t} - \phi_{i}^{t-1} .
    \label{eq:deltaphi}
\end{equation}
What is defined as ``equilibrium'' corresponds here to the value of the similarities between epochs remaining constant, traducing the idea that the $i$-th neuron has learned its input-output relationship. 
To account for the temporal trend of $\Delta \phi_i$, a \emph{velocity} is computed as:
\begin{equation}
    v_{i}^t = \Delta \phi_i^t - \mu_{\text{eq}} v_{i}^{t-1},
    \label{eq:velocity1}
\end{equation}
where $\mu_{\text{eq}}$ is a momentum term, allowing the velocity to carry a memory of its previous values. 

A neuron is then considered \emph{at equilibrium} when its velocity satisfies the following condition:

\begin{equation}
    \left| v_{i}^t \right | < \varepsilon,~~~~~\varepsilon \geq 0. 
    \label{eq:varepscond}
\end{equation}

As the learning process advances (and eventually the learning rate decreases), the neurons' velocities gradually ``slow down'', resulting in more neurons reaching equilibrium. When a neuron is at equilibrium, it can be frozen, allowing for computational savings (as its gradient should not be computed) without hurting the final accuracy (as it has already learned its input-output relationship). In other terms, in the early stage of the training, few neurons are typically frozen as the network is moving to a different loss subspace~\cite{Frankle2020The}. As the training progresses, the network stabilizes: many neurons have learned their target function and do not require further update steps. 

From a very different perspective, NEq proposes an interesting alternative to SU: it does not require any prior knowledge of the target dataset and thus skips the tedious process of analyzing the contributions layers and bias as well as the heavy evolutionary search. Furthermore, it is computed dynamically, selecting the best neurons to update in the network for each epoch. However, it is not off-the-shelf ready to be applied to on-device learning, since in the first epochs many neurons have a high velocity exceeding the threshold value $\varepsilon$, leading to the gradient computation and backpropagation being out of memory.

\subsection{Overhauling the velocity threshold: a budget constraining approach}
\label{sec:NEQ++}

We present here our approach to adapt NEq for enabling effective training complexity reduction satisfying a fixed maximum budget. Intuitively, the online fine-tuning of a network pre-trained on large datasets should endow the neural network with much fewer neurons to update. The pre-training leads the network to learn a latent representation that will be useful for learning the downstream task; thus, as described by Lee~\emph{et~al.}, only a subset of parameters needs to be updated~\cite{lee2022surgical}. We thus replace the $\varepsilon$-threshold freezing condition with a budget for the whole neural network $\mathcal{B}^{w}$, expressed here in terms of the number of updatable parameters. To determine which are the parameters to update, we rank all the neurons in the network along the absolute value of their velocity, from the fastest to the slowest. 
Given $\mathcal{C}_i^{\text{w}}$ the number of parameters of the $i$-th neuron, we evaluate the total parameter's cost including the $j$-th fastest neurons as
\begin{equation}
    \left. \mathcal{B}_j^{\text{w}}  = \sum_{i=1}^j \mathcal{C}_i^w\right|~ v_i \geq v_j,~1 \leq i \leq j \leq M,
\end{equation}
where $M$ is the total number of neurons of the neural network. We will then try to greedily solve the problem of selecting the highest $j$ according to
\begin{equation}
    \label{eq:problem}
    \max_j\{\mathcal{B}_j^{\text{w}} \leq \mathcal{B}_{\text{max}}^{w}\}.
\end{equation}
A graphical representation of the problem in \eqref{eq:problem} is provided in Fig.~\ref{fig:Neuron_selection}. In the figure, the neurons' sizes are proportional to $\mathcal{C}_i^{\text{w}}$. The neurons on the left side of the budget threshold are updated during BP, whereas the neurons on the right are frozen. The budget threshold overlaps a neuron since the cumulative sum grows incrementally with steps of size $\mathcal{C}_i^{\text{w}}$. This way, we can fit a network training under a given budget constraint and dynamically select the best neurons to update at each epoch.

Although \eqref{eq:problem} provides a formulation of the neurons that are changing more rapidly their function, we identify an unfair selection towards neurons having a higher number of parameters. To compensate for this effect, we propose a measure estimating the average velocity-per-parameter 
\begin{equation}
        \tilde{v}_i = \frac{v_i}{\mathcal{C}_i^{\text{w}}} . 
        \label{eq:reweighted_velocity}
\end{equation}
This re-weighted velocity allows us to focus on the neurons with the highest per-parameter average velocity instead of the neurons with the highest global velocity.

\noindent\textbf{A random selection baseline.} 
To the best of our knowledge, our proposal is the very first approach attempting to dynamically select a sub-network to update. A very intuitive baseline we can build consists of randomly selecting neurons to be updated, until the budget $\mathcal{B}_{\text{max}}^{\text{w}}$ is met. 

\begin{table*}
    \caption{Comparison of pretrained MobileNetV2 final top1 test accuracies across different neuron selection methods for three different memory budgets expressed in percentage of network updated and in number of parameters. For the first epoch the neurons to update are given by the associated SU scheme. We highlight with different colors the best method for each budget (blue for Velocity, red for Random and green for SU).}
    \centering
    \renewcommand*{\arraystretch}{1.2}
    \resizebox{\textwidth}{!}{
    \label{tab:results:SU_VS_Velocity}
        \begin{tabular}{cccccccccc}
            \toprule
            \textbf{\% of network updated} & $\mathcal{B}_{\text{max}}^{w}$ & \textbf{Method} & \textbf{Cifar 10} & \textbf{Cifar 100} & \textbf{VWW} & \textbf{Flowers}  & \textbf{Food} & \textbf{Pets} & \textbf{CUB} \\
            \midrule
            \multirow{4}{*}{8.8} & \multirow{4}{*}{192 311} & Sparse Update & 95.13\small$\pm$0.21 & 78.60\small$\pm$0.22 & 90.66\small$\pm$0.29 & \cellcolor{green!15}93.77\small$\pm$0.38 & 77.81\small$\pm$0.26 & \cellcolor{green!15}85.82\small$\pm$0.22 & \cellcolor{green!15}67.82\small$\pm$0.29 \\
    
            \cmidrule{3-10}
            && Velocity & \cellcolor{blue!15}95.25\small$\pm$0.29 & \cellcolor{blue!15}79.46\small$\pm$0.12 & \cellcolor{blue!15}91.40\small$\pm$0.16 & 93.03\small$\pm$0.47 & \cellcolor{blue!15}79.16\small$\pm$0.16 & 85.50\small$\pm$0.17 & 67.52\small$\pm$0.05 \\

            \cmidrule{3-10}
            && Random & 94.41\small$\pm$0.13 & 78.15\small$\pm$0.26 & 90.29\small$\pm$0.05 & 92.19\small$\pm$0.17 & 77.78\small$\pm$0.00 & 85.50\small$\pm$0.28 & 65.56\small$\pm$0.45 \\
            
            \midrule
            \multirow{4}{*}{21.2} & \multirow{4}{*}{464 639} & Sparse Update & 95.30\small$\pm$0.10 & 78.84\small$\pm$0.20 & 91.29\small$\pm$0.18 & \cellcolor{green!15}94.28\small$\pm$0.36 & 78.26\small$\pm$0.07 & 84.63\small$\pm$0.15 & 68.04\small$\pm$0.28 \\
    
            \cmidrule{3-10}
            && Velocity & \cellcolor{blue!15}95.36\small$\pm$0.07 & \cellcolor{blue!15}79.67\small$\pm$0.28 & \cellcolor{blue!15}91.48\small$\pm$0.39 & 93.34\small$\pm$0.08 & \cellcolor{blue!15}79.63\small$\pm$0.17 & \cellcolor{blue!15}84.91\small$\pm$0.82 & \cellcolor{blue!15}68.23\small$\pm$0.61 \\

            \cmidrule{3-10}
            && Random & 94.61\small$\pm$0.16 & 78.28\small$\pm$0.31 & 90.51\small$\pm$0.25 & 92.43\small$\pm$0.10 & 78.26\small$\pm$0.07 & 84.73\small$\pm$0.29 & 66.30\small$\pm$0.13 \\
            
            \midrule
            \multirow{4}{*}{30.8} & \multirow{4}{*}{675 540} & Sparse Update & 95.16\small$\pm$0.29 & 78.62\small$\pm$0.18 & 91.46\small$\pm$0.21 & \cellcolor{green!15}94.22\small$\pm$0.14 & 78.03\small$\pm$0.08 & 84.38\small$\pm$0.28 & 67.59\small$\pm$0.22 \\
    
            \cmidrule{3-10}
            && Velocity & \cellcolor{blue!15}95.49\small$\pm$0.16 & \cellcolor{blue!15}79.43\small$\pm$0.19 & \cellcolor{blue!15}91.57\small$\pm$0.20 & 93.77\small$\pm$0.26 & \cellcolor{blue!15}79.56\small$\pm$0.24 & \cellcolor{blue!15}84.44\small$\pm$0.50 & \cellcolor{blue!15}68.26\small$\pm$0.36 \\

            \cmidrule{3-10}
            && Random & 94.57\small$\pm$0.20 & 78.58\small$\pm$0.11 & 90.58\small$\pm$0.10 & 92.83\small$\pm$0.03 & 79.00\small$\pm$0.11 & 84.39\small$\pm$0.42 & 66.17\small$\pm$0.42 \\
            
            \bottomrule
            
        \end{tabular}
    }
\end{table*}

\subsection{Training algorithm for a dynamic neuron's update selection strategy}

\begin{figure}[t]
    \centering
     \includegraphics[width=\columnwidth]{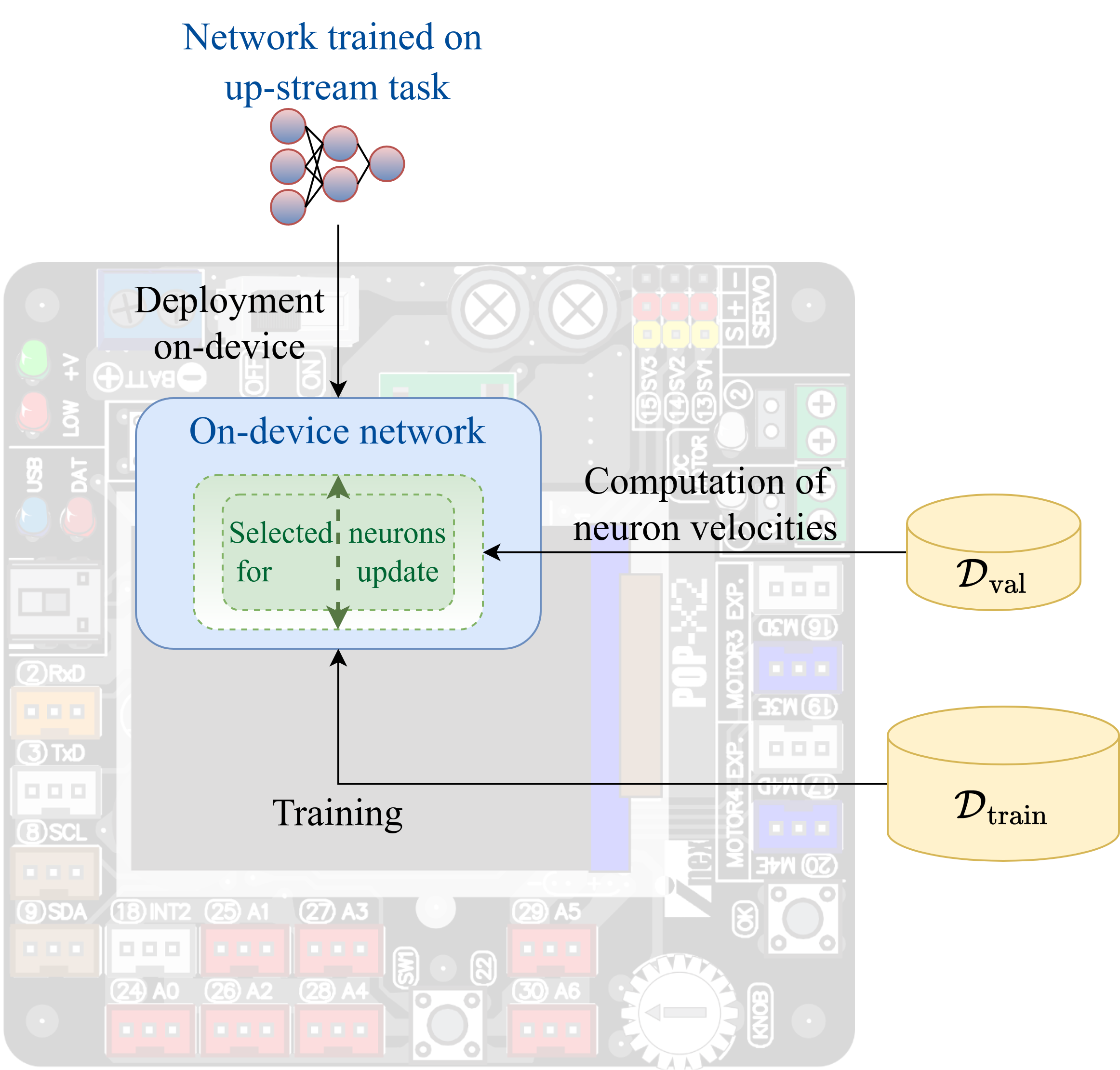}
    \caption{Overview for the on-device learning when dynamic neuron selection is applied.} 
    \label{fig:dynamic_recap}\vspace{-3ex}
\end{figure}

To summarize, the on-device learning strategies all involve pre-training a model on some upstream task, which is then loaded to the target device. Subsequently, either a static or a dynamic neuron selection strategy can take place - for a neuron velocity-based selection the use of a validation set $\mathcal{D}_{\text{val}}$ is required - and training on the $\mathcal{D}_{\text{train}}$ data of the target down-stream task takes place (as synthetically visualized in Fig.~\ref{fig:dynamic_recap}). Overall, the neuron selection strategy for on-device fine-tuning is composed of three distinct phases.
\begin{enumerate}
    \item Select the sub-network to be updated for the first epoch: this phase can leverage knowledge from an evolutionary search algorithm as in SU, or in a random selection for the subgraph to update, in compliance with the maximum budget $\mathcal{B}_{\text{max}}^{w}$.
    \item Train for one epoch.
    \item Evaluate which part of the whole model should be updated for the next epoch. The sub-graph of neurons to be updated will not change when using SU, it can change when employing either a velocity-based selection or a random selection. The training is iterated back to step 2 until some target training termination conditions are met (in our case, a wall number of epochs).
\end{enumerate}
This very general framework allows for plugging any dynamic neuron selection strategy, and we will use it to compare static neuron selection (SU), velocity-based selection, and random selection. In the next section, we will present some results obtained on popular benchmarks.
\section{Experiments}
\begin{figure}[t]
    \centering
    \includegraphics[width=0.85\columnwidth]{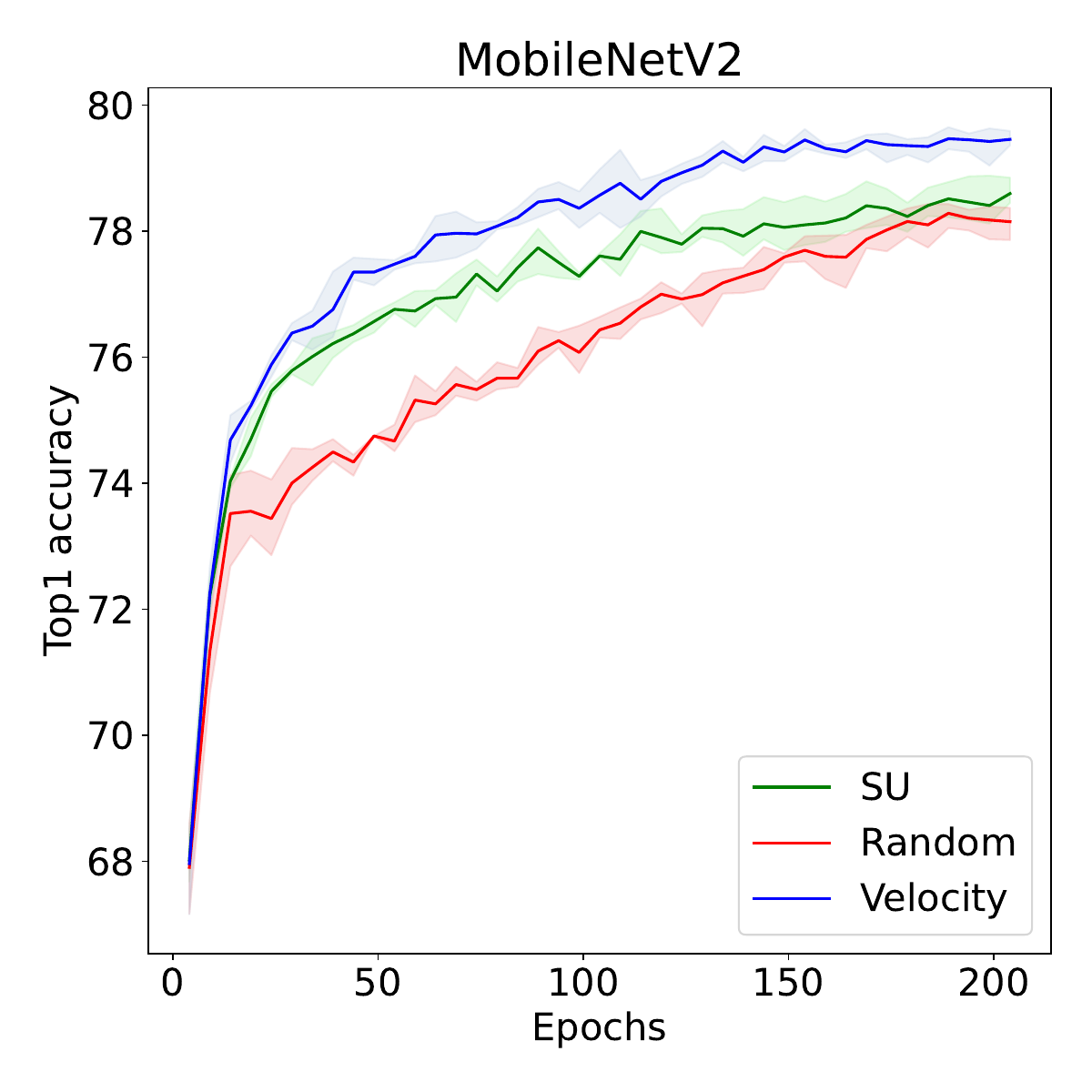}
    \caption{Comparison of MobileNetV2 testing accuracy through training on C100. We update 8.8\% of the network's parameters.}
    \label{fig:mbv2_acc}
\end{figure}
\begin{table*}
    \caption{Comparison of final top1 test accuracies between Baseline, Sparse Update (SU), Random and Velocity neuron selection over various pretrained models, datasets, and budgets. For the first epoch the neurons to update are randomly selected. For each budget, highlighted results correspond to the best accuracy between neuron selection methods (green if SU is better, blue for Velocity, and red for Random, in bold the overall best performance regardless of the budget).}
    \centering
    \renewcommand*{\arraystretch}{1.2}
    \resizebox{\textwidth}{!}{
    \label{tab:results:Velocity_VS_Random}
        \begin{tabular}{cccccccccc}
            \toprule
            \textbf{Model} & $\mathcal{B}_{\text{max}}^{w}$ &\textbf{Method} & \textbf{Cifar 10} & \textbf{Cifar 100} & \textbf{VWW} & \textbf{Flowers}  & \textbf{Food} & \textbf{Pets} & \textbf{CUB} \\
            
            \midrule
            \multirow{13}{*}{MbV2} 
            & \multirow{4}{*}{192 311} & SU & 94.88\small$\pm$0.12 & 78.15\small$\pm$0.13 & 90.75\small$\pm$0.17 & 92.70\small$\pm$0.06 & 75.02\small$\pm$0.23 & \bf \cellcolor{green!15}86.93\small$\pm$0.22 & 66.48\small$\pm$0.41 \\

            \cmidrule{3-10}
            & & Velocity & \cellcolor{blue!15}95.35\small$\pm$0.35 & \cellcolor{blue!15}79.41\small$\pm$0.21 & \cellcolor{blue!15}90.95\small$\pm$0.16 & \cellcolor{blue!15}92.98\small$\pm$0.29 & \cellcolor{blue!15}79.18\small$\pm$0.07 & 85.56\small$\pm$0.47 & \cellcolor{blue!15}67.92\small$\pm$0.27 \\

            \cmidrule{3-10}
            & & Random & 94.46\small$\pm$0.16 & 78.03\small$\pm$0.21 & 90.20\small$\pm$0.13 & 92.11\small$\pm$0.15 & 77.57\small$\pm$0.16 & 85.16\small$\pm$0.07 & 65.96\small$\pm$0.06 \\

            \cmidrule{2-10}
            & \multirow{4}{*}{464 639} & SU & 95.00\small$\pm$0.08 & 78.69\small$\pm$0.19 & 90.80\small$\pm$0.24 & 92.86\small$\pm$0.33 & 76.50\small$\pm$0.23 & \cellcolor{green!15}86.64\small$\pm$0.27 & 67.81\small$\pm$0.23 \\
            
            \cmidrule{3-10}
            & & Velocity & \cellcolor{blue!15}95.49\small$\pm$0.02 & \cellcolor{blue!15}79.52\small$\pm$0.11 & \cellcolor{blue!15}91.41\small$\pm$0.19 & \cellcolor{blue!15}93.32\small$\pm$0.15 & \cellcolor{blue!15}79.67\small$\pm$0.19 & 84.36\small$\pm$0.76 & \cellcolor{blue!15}68.19\small$\pm$0.24 \\

            \cmidrule{3-10}
            & & Random & 94.51\small$\pm$0.05 & 78.49\small$\pm$0.19 & 90.55\small$\pm$0.24 & 92.64\small$\pm$0.32 & 78.48\small$\pm$0.17 & 84.92\small$\pm$0.21 & 66.18\small$\pm$0.71 \\

            \cmidrule{2-10}
            & \multirow{4}{*}{675 540} & SU & 95.18\small$\pm$0.17 & 79.03\small$\pm$0.30 & 91.03\small$\pm$0.16 & 93.08\small$\pm$0.09 & 77.19\small$\pm$0.07 & \cellcolor{green!15}86.42\small$\pm$0.45 & 67.72\small$\pm$0.32 \\

            \cmidrule{3-10}
            & & Velocity & \cellcolor{blue!15}95.57\small$\pm$0.11 & \cellcolor{blue!15}79.21\small$\pm$0.37 & \cellcolor{blue!15}91.72\small$\pm$0.15 & \cellcolor{blue!15}93.33\small$\pm$0.31 & \cellcolor{blue!15}79.68\small$\pm$0.16 & 84.37\small$\pm$0.28 & \cellcolor{blue!15}68.19\small$\pm$0.24 \\

            \cmidrule{3-10}
            & & Random & 94.57\small$\pm$0.09 & 78.41\small$\pm$0.29 & 90.35\small$\pm$0.01 & 92.88\small$\pm$0.21 & 78.90\small$\pm$0.06 & 84.39\small$\pm$0.41 & 66.36\small$\pm$0.20 \\

            \cmidrule{2-10}
            & 2 189 760 & Baseline & \bf 95.93\small$\pm$0.14 & \bf 79.83\small$\pm$0.29 & \bf 91.80\small$\pm$0.03 & \bf 94.02\small$\pm$0.03 & \bf 80.63\small$\pm$0.10 & 82.82\small$\pm$0.18 & \bf 69.24\small$\pm$0.34 \\
            \midrule

            \multirow{9}{*}{Resnet18} & \multirow{2.5}{*}{980 715} & Velocity & \cellcolor{blue!15}95.51\small$\pm$0.10 & \cellcolor{blue!15}78.77\small$\pm$0.41 & \cellcolor{blue!15}88.78\small$\pm$0.51 & \cellcolor{blue!15}90.78\small$\pm$0.24 & \cellcolor{blue!15}75.09\small$\pm$0.13 & \bf \cellcolor{blue!15}82.82\small$\pm$0.30 & \cellcolor{blue!15}63.64\small$\pm$0.35 \\
            
            \cmidrule{3-10}
            & & Random & 95.20\small$\pm$0.20 & 77.98\small$\pm$0.38 & 88.33\small$\pm$0.45 & 89.39\small$\pm$0.47 & 74.57\small$\pm$0.15 & 79.49\small$\pm$0.51 & 60.93\small$\pm$0.54 \\

            \cmidrule{2-10}
            & \multirow{2.5}{*}{2 369 480} & Velocity & 95.36\small$\pm$0.15 & \bf \cellcolor{blue!15}79.12\small$\pm$0.12 & 89.16\small$\pm$0.31 & \bf \cellcolor{blue!15}91.02\small$\pm$0.17 & \cellcolor{blue!15}75.72\small$\pm$0.23 & \cellcolor{blue!15}82.01\small$\pm$0.60 & \bf \cellcolor{blue!15}63.84\small$\pm$0.39 \\

            \cmidrule{3-10}
            & & Random & \cellcolor{red!25}95.68\small$\pm$0.10 & 78.28\small$\pm$0.22 & \cellcolor{red!25}89.21\small$\pm$0.23 & 89.53\small$\pm$0.17 & 75.17\small$\pm$0.12 & 79.40\small$\pm$0.34 & 61.42\small$\pm$0.32 \\

            \cmidrule{2-10}
            & \multirow{2.5}{*}{3 444 987} & Velocity & 95.58\small$\pm$0.21 & \cellcolor{blue!15}78.95\small$\pm$0.13 & \cellcolor{blue!15}89.17\small$\pm$0.33 & \cellcolor{blue!15}90.76\small$\pm$0.15 & \cellcolor{blue!15}75.83\small$\pm$0.12 & \cellcolor{blue!15}81.45\small$\pm$0.63 & \cellcolor{blue!15}63.59\small$\pm$0.03 \\

            \cmidrule{3-10}
            & & Random & \cellcolor{red!25}95.80\small$\pm$0.09 & 78.52\small$\pm$0.18 & 88.92\small$\pm$0.19 & 89.66\small$\pm$0.30 & 75.28\small$\pm$0.12 & 79.28\small$\pm$0.34 & 61.45\small$\pm$0.32 \\

            \cmidrule{2-10}
            & 11 166 912 & Baseline & \bf 96.2\small$\pm$0.13 & 78.86\small$\pm$0.08 & \bf 89.78\small$\pm$0.24 & 90.14\small$\pm$0.26 & \bf 76.32\small$\pm$0.08 & 79.76\small$\pm$0.63 & 60.97\small$\pm$0.52 \\

            \midrule
            \multirow{9}{*}{Resnet50} & \multirow{2.5}{*}{2 059 888} & Velocity & \cellcolor{blue!15}97.10\small$\pm$0.07 & \bf \cellcolor{blue!15}82.94\small$\pm$0.35 & \cellcolor{blue!15}93.04\small$\pm$0.15 & 93.65\small$\pm$0.08 & \cellcolor{blue!15}81.10\small$\pm$0.05 & \bf \cellcolor{blue!15}90.11\small$\pm$0.25 & \bf \cellcolor{blue!15}73.73\small$\pm$0.52 \\

            \cmidrule{3-10}
            & & Random & 96.80\small$\pm$0.06 & 81.46\small$\pm$0.11 & 92.13\small$\pm$0.33 & \cellcolor{red!25}94.04\small$\pm$0.18 & 80.68\small$\pm$0.18 & 88.92\small$\pm$0.18 & 72.79\small$\pm$0.15 \\

            \cmidrule{2-10}
            & \multirow{2.5}{*}{4 976 842} & Velocity & \cellcolor{blue!15}97.12\small$\pm$0.09 & \cellcolor{blue!15}82.79\small$\pm$0.40 & \bf \cellcolor{blue!15}93.37\small$\pm$0.14 & \cellcolor{blue!15}94.84\small$\pm$0.08 & \cellcolor{blue!15}81.62\small$\pm$0.17 & \cellcolor{blue!15}89.42\small$\pm$0.25 & \cellcolor{blue!15}73.55\small$\pm$0.18 \\

            \cmidrule{3-10}
            & & Random & 96.97\small$\pm$0.08 & 82.04\small$\pm$0.11 & 92.59\small$\pm$0.20 & 94.23\small$\pm$0.22 & 81.52\small$\pm$0.11 & 88.46\small$\pm$0.07 & 72.40\small$\pm$0.60 \\

            \cmidrule{2-10}
            & \multirow{2.5}{*}{7 235 830} & Velocity & \cellcolor{blue!15}97.07\small$\pm$0.08 & \cellcolor{blue!15}82.45\small$\pm$0.18 & \cellcolor{blue!15}93.21\small$\pm$0.14 & \bf \cellcolor{blue!15}95.03\small$\pm$0.18 & 81.76\small$\pm$0.06 & \cellcolor{blue!15}88.97\small$\pm$0.60 & \cellcolor{blue!15}73.54\small$\pm$0.42 \\
            
            \cmidrule{3-10}
            & & Random & 97.01\small$\pm$0.01 & 82.27\small$\pm$0.29 & 92.59\small$\pm$0.21 & 94.54\small$\pm$0.15 & \cellcolor{red!25}81.88\small$\pm$0.29 & 88.18\small$\pm$0.63 & 72.40\small$\pm$0.35\\

            \cmidrule{2-10}
            & 23 454 912 & Baseline & \bf 97.30\small$\pm$0.04 & 82.63\small$\pm$0.25 & 92.91\small$\pm$0.22 & 94.87\small$\pm$0.20 & \bf 82.49\small$\pm$0.15 & 87.29\small$\pm$0.34 & 72.93\small$\pm$0.41 \\

            \bottomrule            
        \end{tabular}
    }
\end{table*}
In this section, we will describe the experiments conducted to validate our hypotheses. We selected 7 target datasets to fine-tune our models upon, mostly to compare our results with SU performances as these are the datasets they selected in their study. The datasets are Cifar10, Cifar100, CUB~\cite{wah2011caltech}, Flowers~\cite{Nilsback08}, Food~\cite{bossard2014food}, Pets~\cite{zhang20220} and Visual Wake Words~\cite{chowdhery2019visual}. We selected three models to test on (MobileNetV2, Resnet18, and Resnet50) and we averaged the results over 3 different seeds. Our fine-tuning policy is the same for all the experiments: we use SGD with no momentum and no weight decay and a cosine annealing scheduler over 200 epochs, ranging from $0.125$ to $0$ with 5 warm-up epochs. We report the final top1 accuracy obtained on the test set $\mathcal{D}_{\text{test}}$. All the experiments were conducted on an Nvidia RTX3090Ti with 24GB, algorithms are implemented in Python, using PyTorch 2.0.0.\footnote{The code used to run experiments is available at https://github.com/aelQuelennec/-WACV-2024-Ways-to-Select-Neurons-under-a-Budget-Constraint.}

\subsection{Dynamic neuron selection VS SU}
\label{sec:dynvsstatic}

Our first set of experiments is the comparison of performance between SU and dynamic neuron selection methods. We isolated the SU code from the other technologies described in the original article and compared the results of fine-tuning with Velocity and Random (Table~\ref{tab:results:SU_VS_Velocity}). We tested on MobileNetV2 with three different levels of budget constraint provided by the exact SU configurations Lin~\emph{et~al.} designed for the network~\cite{lin2022device}. We calculate the number of parameters updated by a given SU scheme and we use that number as a budget for Velocity and Random. It is important to note that for the first epoch, we can not compute the neurons' velocities: we thus use the SU scheme to compare final results given an identical starting point. We observe that in the large majority of the cases, the update based on Velocity is competitive with SU and outperforms the random update baseline. This confirms that a proper selection of neurons to update guides the model to better generalizing solutions.

In Fig.~\ref{fig:mbv2_acc}, we display the test accuracy evolution for the different neuron selections. From this, we clearly observe that Velocity approaches faster than other approaches with high generalization. As intuitive, SU plateaus at intermediate epochs - seemingly, the updated parameters land at a local minimum. The Random update scheme is in general the worst, but progressively is able to catch up with the other approaches.

\begin{figure*}[t]
    \begin{subfigure}{0.33\textwidth}
        \includegraphics[width=0.9\linewidth]{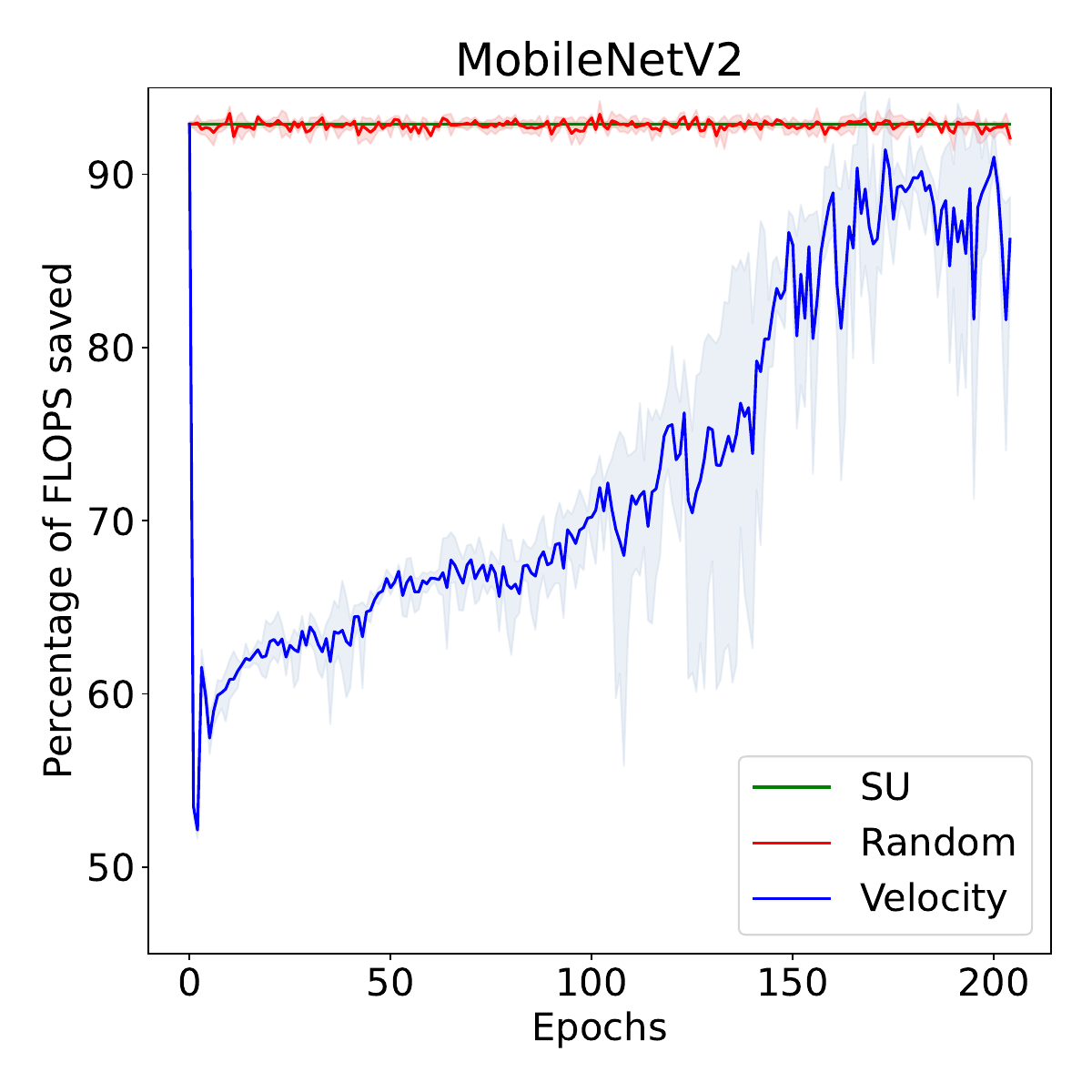} 
        \label{fig:subimMBV2}
    \end{subfigure}
    \begin{subfigure}{0.33\textwidth}
        \includegraphics[width=0.9\linewidth]{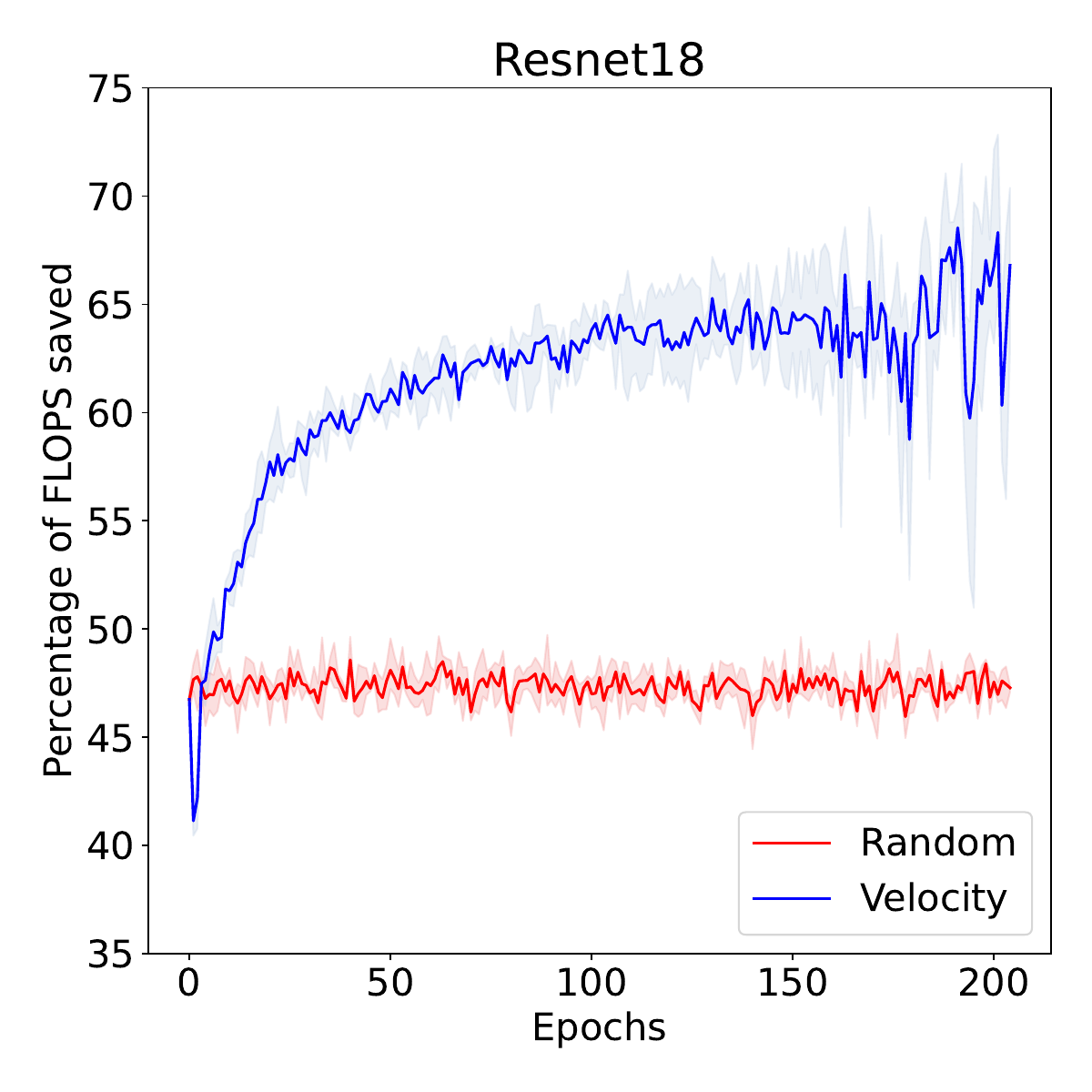}
        \label{fig:subimR18}
    \end{subfigure}
    \begin{subfigure}{0.33\textwidth}
        \includegraphics[width=0.9\linewidth]{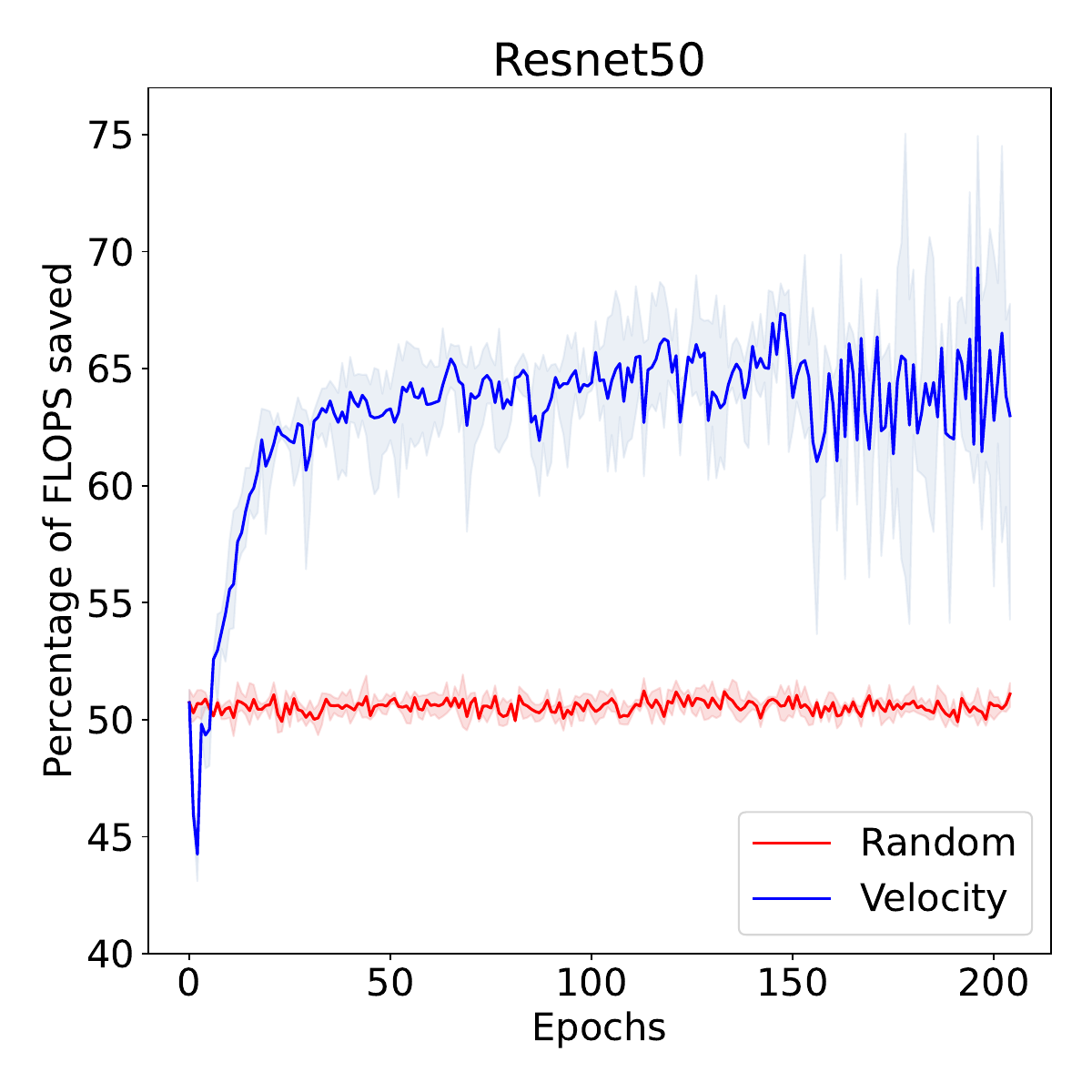}
        \label{fig:subimR50}
    \end{subfigure}
    
    \caption{Percentage of FLOPS saved with SU, Velocity and Random neuron selection for each network during Cifar100 fine-tuning. We update 30.2\% of the network's parameters.}
    \label{fig:Flops_savings}
\end{figure*}

\subsection{Velocity VS Random}
\label{sec:dynamicvsrandom}
In the attempt to assess a comparison between Velocity and Random neuron selection methods, we ran multiple experiments on the various models and datasets selected. Here the budgets we test are deduced from the percentage of total parameters updated in the SU schemes (on MobilenetV2) and applied to the Resnets architectures. For the first epoch, since we do not dispose of the neurons' velocities, we initialize the neurons randomly. The results obtained are displayed in Table~\ref{tab:results:Velocity_VS_Random}. Comparing the experiments on Velocity with the SU initialization proposed in Table~\ref{tab:results:SU_VS_Velocity}, we observe little to no impact on final accuracy for either of the two initialization methods. This suggests that we can dispense from finding the SU schemes and instead rely on dynamic neuron selection throughout the whole training. More broadly, we observe that MobileNetV2 clearly favors the Velocity selection (except in the case of the Pets dataset), whereas in Resnet networks the Random selection can result in marginally better performance, in some rare cases. It is important to note that as the budgets are computed as identical percentages of the network, they are tighter in the absolute value of parameters updated for the MobileNetV2 than for the Resnets. 

\noindent\textbf{Gap with the baseline models.} Besides the comparisons between the Random and the Velocity approaches, in Table~\ref{tab:results:Velocity_VS_Random} we also report the baseline performance, when updating the full network. This gives us a reference accuracy (virtually achievable). We do not observe a big gap in performance between the low-budget approaches and the full model's update, and in some cases, the reported performance is even higher than the baseline, like in Pets for MobileNetV2 and ResNet50, where we observe an improvement of approximately 3 to 4\%. We can explain this effect by the reduced size of the datasets, and the limited budget of parameters employed to fine-tune: we implicitly avoid overfitting.

\noindent\textbf{FLOPs saved at training time.} To qualitatively evaluate the impact on the training complexity for either of the two approaches, we introduce here as an efficiency metric the percentage of FLOPS saved (in comparison with the full model training). 
We observe that Random is a more economical strategy (more FLOPS are saved) for MobileNetV2, whereas it is the opposite for Resnets, as displayed in Fig.~\ref{fig:Flops_savings}). 
While controlling FLOPs savings for Random is relatively straightforward, it becomes less manageable for a dynamic strategy like Velocity. This poses a critical consideration when implementing a learning strategy on the edge. This highlights the need for the development of new selection methods that also account for energy consumption, which remains a topic for future exploration.

\section{Conclusions and Future works}

In this work, two resource-saving neural update philosophies were presented: a static strategy, proposed in~\cite{lin2022device}, where a sub-network in a pre-trained model is identified prior to training on the device (and statically determined for any downstream tasks), and a dynamic strategy. The results obtained suggest that the dynamic selection of neurons can be better than the static pre-selection of a sub-network to train. Specifically, the proposed dynamic strategy, which is an evolution of a resources-unconstrained training strategy~\cite{bragagnolo2022update}, proved effective in almost all the scenarios. 

Other aspects to be considered for progress in this field are the inclusion of activation cost and (expected) training cost for the selected neurons, and considering strategies potentially saving computation at forward-propagation time too, such as dropout~\cite{srivastava2014dropout}.

\section*{Acknowledgements}
This research was partially funded by Hi!PARIS Center on Data Analytics and Artificial Intelligence and by the project titled “PC2-NF-NAI” in the
framework of the program “PEPR 5G et Réseaux du futur”. This paper is supported by the European Union’s HORIZON Research and Innovation Programme under grant agreement No 101120657, project ENFIELD (European Lighthouse to Manifest Trustworthy and Green AI). 
{\small
\bibliographystyle{ieee_fullname}
\bibliography{egbib}
}

\end{document}